# Conceptual Knowledge Markup Language: An Introduction

**Robert E. Kent**

**School of EECS**

**Washington State University**

## Abstract

*Conceptual Knowledge Markup Language (CKML) is an application of [XML](). Earlier versions of CKML followed rather exclusively the philosophy of [Conceptual Knowledge Processing (CKP)](), a principled approach to knowledge representation and data analysis that "advocates methods and instruments of conceptual knowledge processing which support people in their rational thinking, judgment and acting and promote critical discussion." The new version of CKML continues to follow this approach, but also incorporates various principles, insights and techniques from [Information Flow (IF)](), the logical design of distributed systems. Among other things, this allows diverse communities of discourse to compare their own information structures, as coded in logical theories, with that of other communities that share a common generic ontology. CKML incorporates the CKP ideas of concept lattice and formal context, along with the IF ideas of classification (= formal context), informorphism, theory, interpretation and local logic. Ontology Markup Language (OML), a subset of CKML that is a self-sufficient markup language in its own right, follows the principles and ideas of [Conceptual Graphs (CG)](). OML is used for structuring the specifications and axiomatics of metadata into ontologies. OML incorporates the CG ideas of concept, conceptual relation, conceptual graph, conceptual context, participants and ontology. The link from OML to CKML is the process of conceptual scaling, which is the interpretive transformation of ontologically structured knowledge to conceptual structured knowledge.*

**Diagram 1: OML/CKML Architecture**

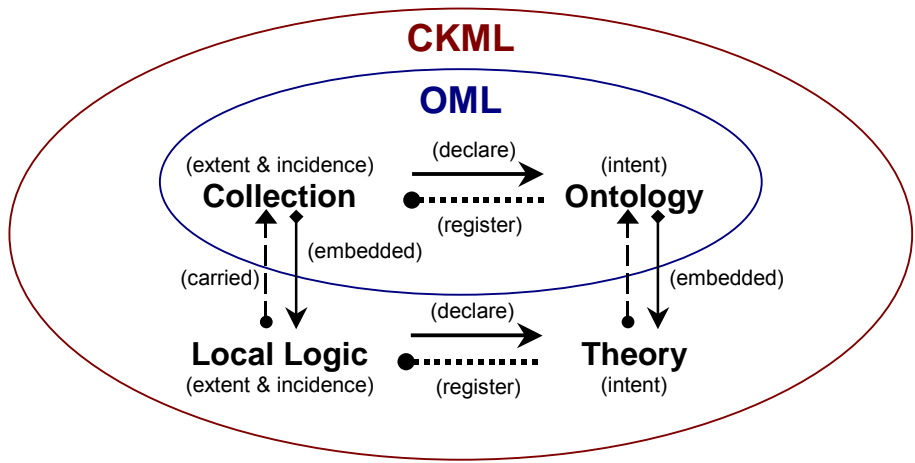

## Introduction

Conceptual Knowledge Markup Language (CKML) provides a specification standard for the conceptual representation and analysis of networked resources. Although hypertext links enable organization in the small, resource discovery systems, using content-based access to documents, enable organization in the large [2]. Content-based access requires a good representation for document content and the natural hierarchies associated with documents and related entities [5]. Ideally, such document content and natural hierarchies would be transparently composable. Concept lattices allow such transparent composability by representing both content and entity type hierarchies as facets of document information.

A *concept lattice* is a lattice with bound objects and attributes. Any arbitrary lattice is a concept lattice, where the objects and attributes bound to it are defined mathematically - they are the atomic (irreducible) elements with respect to joins and meets. Object-attribute incidence relations, called formal contexts (CKP) or classifications (IF), and concept lattices are equivalent structures. Either provides for the conceptual representation of networked resources. When applying conceptual knowledge processing to networked resources, objects are represented as abstracted ontologically structured metadata and attributes are defined by logical queries. The process of conceptual scaling constructs concept lattices, and their equivalent formal contexts. Conceptual scaling transforms ontologically structured collections of objects to faceted conceptual space by applying conceptual scales. Figure 1 illustrates conversion of the intuition behind the ontological knowledge model into the formalism of CKML.

**Figure 1: From Ontological Intuition to Conceptual Formalism**

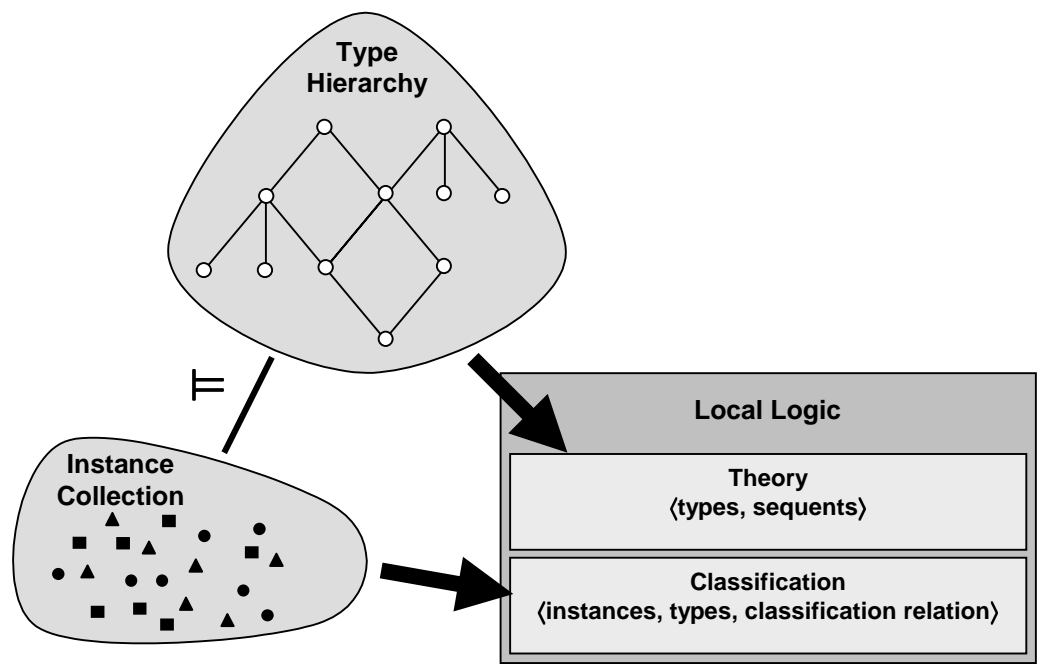

The goal of this paper is to introduce the reader to CKML. The first section on the CKML knowledge model covers the semantics of CKML. The discussion of semantics ranges over ontologies and conceptual knowledge processing. Illustrative examples also introduce the syntax of CKML. In the second section the pragmatics of CKML is made with a real world example - the online press releases of Intel Corporation. The third section on CKML interoperability is an argument for the position that CKML provides a generic solution for structuring online information. The final section discusses a recent and extensive evaluation of the CKML language for use as an interchange language in the life sciences community.

# The CKML Knowledge Model

## OML/CKML AT A GLANCE

As justified by the following points, CKML is in various senses both a description logic based language and a frame-based language.

- **CKML:** Being based upon conceptual graphs, formal concept analysis, and information flow, the CKML extension of OML is closely related to a description logic based approach for modeling ontologies. Conceptual scaling and concept lattice algorithms correspond to subsumption.
- **OML:** How and how well a knowledge representation language expresses constraints is a very important issue. OML has three levels for constraint expression:
  - top – sequents
  - intermediate – calculus of binary relations
  - bottom – logical expressions

  The top level models the theory constraints of information flow, the middle level arises both from the practical importance of binary relation constraints and the category theory orientation to the classification-projection semantics of Diagram 3, and the bottom level corresponds to conceptual graphs. Indeed, OML assertions (closed expressions) correspond exactly to conceptual graphs.
- **Simple OML:** The semantic core of CKML is captured in a subset of OML called Simple OML. Because Simple OML is based upon the fundamental classification-projection semantics of Diagram 3, the normal expression of types and instances is very frame-like. A core aspect of Open Knowledge Base Connectivity (OKBC) is closely related to Simple OML.

## ONTOLOGIES

Ontologies provide a knowledge model for expressing the structured metadata of both concrete and abstract distributed objects. They have proven to be an excellent basis on which to build facilities for conceptual knowledge processing. As such, the ontology-based knowledge models are well suited to the goal of CKML to provide a foundational framework for the specification of conceptual knowledge. In facet, ontologically structured information has been used as a challenge to OML/CKML during its continuing development. This includes ontologies for: corporations, movies, television, gene function in molecular biology, medical informatics, etc. Most of these are accessible through the links at the site ontologos.org.

The notion of ontology is widely used in the fields of knowledge representation and artificial intelligence. The notion of ontology used in OML/CKML closely follows Gruber's definition that "an ontology is an explicit specification of a conceptualization." In CKML the conceptualization being specified is the mathematical notion of a concept lattice or the more practical notion of a conceptual space. An ontological specification consists of a set of entities structured a taxonomic hierarchy of types, relationships between those entities, and appropriate semantic constraints involving those entities and relationships. The taxonomic hierarchy is a key feature of an ontology. As Luke [7] asserts, "The ability to establish relationships is important, but secondary to the ability to classify those entities." The important conceptual scaling link from OML to CKML converts both static taxonomic information and ontological relational information to the dynamic hierarchical structure of the concept lattice.

## CONCEPTUAL KNOWLEDGE REPRESENTATION

The purpose for CKML is to provide a mechanism whereby people can conceptually organized their information. This conceptual approach to information organization is based upon certain principles and ideas from conceptual knowledge processing. The central structuring methodology in conceptual knowledge processing, called a *concept lattice* [11], is a lattice with objects and attributes bound to its atomic (join or meet irreducible) elements. Elements of the concept lattice, which can be multiply labeled with objects or attributes, are called *formal concepts*.

Any formal concept is identified with the collection of objects below it in the lattice called its extent, and the collection of attributes above it called its intent. The intent of a formal concept contains exactly the attributes that are possessed by all objects in the extent, and vice-versa the extent of a formal concept contains exactly the objects that possess all the attributes in the intent. A concept $c_1$ is below another concept $c_2$ in the lattice when the extent of $c_1$ is contained in the extent of $c_2$, or dually the intent of $c_1$ contains the intent of $c_2$, indicating that $c_1$ is more specialized than $c_2$. We calculate the meet of two concepts by intersecting their extents (and then computing the resulting intent). Dually, we calculate the join of two concepts by intersecting their intents (and then computing the resulting extent).

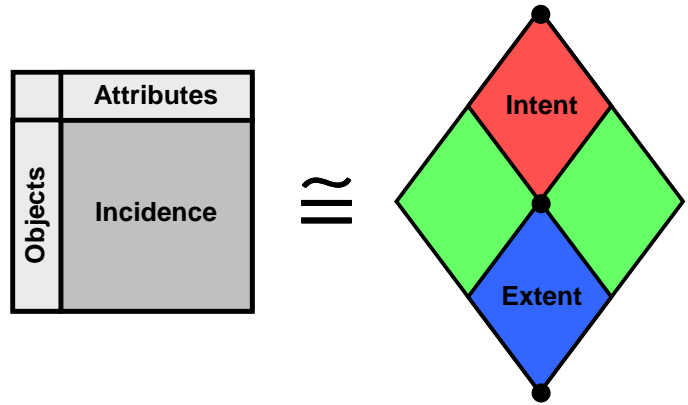

Figure 2 illustrates the basic theorem of conceptual knowledge processing [11], that formal contexts (classifications) and concept lattices are equivalent data structures. The rectangular component in Figure 2 represents a formal context, whereas the diamond-shaped component represents a concept lattice. The central node within the concept lattice of Figure 2 represents an arbitrary formal concept. A *formal context* (or *classification*) is essentially an incidence relation (matrix) between a set of objects and a set of (single-valued) attributes. If we represent the fact "object $o$ in row $i$ has attribute $a$ in column $j$" by a boolean "1" at entry $(i, j)$, with "0"s otherwise, then a formal concept in the associated concept lattice corresponds to a maximal rectangle of "1"s in the matrix (modulo change-of-basis permutations of either objects or attributes) with sides of the rectangle indexed by extent and intent.

## Example: The Living Concept Space

The "living" concept space is a tiny dataset, which exists within a conceptual universe of living organisms. This space consists of eight organisms (plants and animals), and nine of their properties. The organisms are the *objects* of the space, and the properties are the *attributes*.

### Objects

| Le | Leech |
|----|-------|
| Br | Bream |
| Fr | Frog |
| Dg | Dog |
| SW | Spike-Weed |
| Rd | Reed |
| Bn | Bean |
| Ma | Maize |

### Attributes

| nw | needs water |
|----|-------------|
| lw | lives in water |
| ll | lives on land |
| nc | needs chlorophyll |
| 2lg | 2 leaf germination |
| 1lg | 1 leaf germination |
| mo | is motile |
| lb | has limbs |
| sk | suckles young |

The living concept space is illustrated in Diagram 2. On the left is the classification (formal context) between objects and attributes. The attributes of conceptual knowledge processing correspond to the types of information flow. On the right is the concept lattice with 19 formal concepts.

**Diagram 2: The Living Concept Space**

**classification (formal context)**　　　　　　　　　**concept lattice**

| | nw | lw | ll | nc | 2lg | 1lg | mo | lb | sk |
|---|---|---|---|---|---|---|---|---|---|
| Le | × | × | | | | | × | | |
| Br | | | | | | | × | × | |
| Fr | × | × | | | | | × | × | |
| Dg | × | | × | | | | × | × | × |
| SW | × | × | | × | | × | | | |
| Rd | × | × | × | × | | × | | | |
| Bn | × | | | × | × | | | | |
| Ma | × | | × | × | | × | | | |

$\cong$

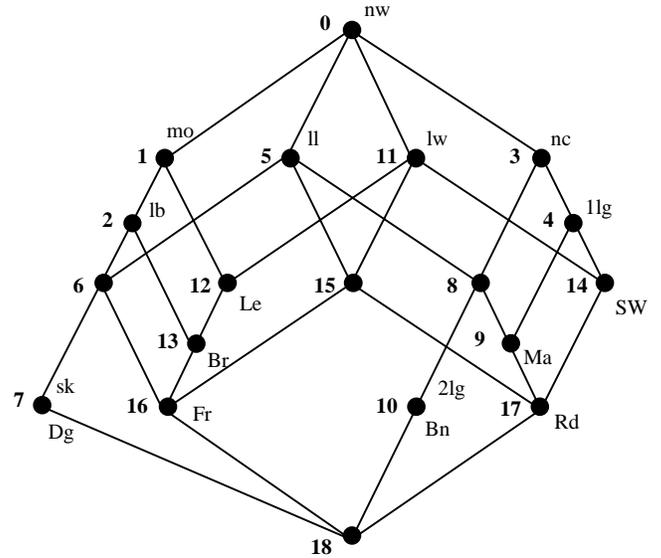

Two of those formal concepts, along with their type ordering in the lattice, are described below.

> concept **#9** = ⟨{Reed, Maize}, {needs water, lives on land, needs chlorophyll, 1 leaf germination}⟩
> concept **#4** = ⟨{Spike-Weed, Reed, Maize}, {needs water, needs chlorophyll, 1 leaf germination}⟩
> concept **#9** ≤ concept **#4**

The structure of a concept lattice can be specified with theory constraints. A *theory constraint* (sequent) is a pair of type sets $\Gamma \vdash \Delta$ with the meaning "the conjunction (**and**) of the types in $\Gamma$ implies the disjunction (**or**) of the types in $\Delta$." Theory constraints allow the specification of disjointness, subtype-supertype and covering relationships.

- **disjoint concepts (types)**

| traditional formalism | sequent notation |
|---|---|
| nc ⊥ mo | nc, mo ⊢ |
| 2lg ⊥ 1lg | 2lg, 1lg ⊢ |

- **subconcepts**

| traditional formalism | sequent notation |
|---|---|
| sk ≤ lb ≤ mo | sk ⊢ lb ⊢ mo |
| 1lg ≤ nc | 1lg ⊢ nc |
| **any** ≤ nw | **any** ⊢ nw |

- **covers**

| traditional formalism | sequent notation |
|---|---|
| ll ⎪ mo | ⊢ ll, mo |
| mo ⎪ nc | ⊢ mo, nc |

The CKML expression for disjointness follows closely the sequent notation. The fact " 'Needs Chlorophyll' and 'Is Motile' are disjoint (nc, mo ⊢) is expressed as follows.

```
<sequent>
  <li type="Needs Chlorophyll"/>
  <li type="Is Motile"/>
<entails/>
</sequent>
```

The CKML expression for the subtype-supertype relationship is naturally expressed. The fact that " 'Has Limbs' is a subtype of 'Is Motile' " is expressed as follows.

```
<subtype specific="Has Limbs" generic="Is Motile"/>
```

Subtyping, disjointness and covering define partition. Partition is also naturally expressed in CKML. The fact that " The types 'Needs Chlorophyll' and 'Is Motile' partition the root type 'Needs Water' " is expressed in CKML as follows.

```
<partition genus="Needs Water">
  <li type="Needs Chlorophyll"/>
  <li type="Is Motile"/>
</partition>
```

The sequent notation is primitive, and other notations such as subtype and partition can be expressed in terms of the sequent notation. For example, the partition above can be expressed in CKML sequent notation as follows. Obviously, partition is a good and meaningful abbreviation.

```
<sequent><li type="Needs Chlorophyll"/><entails/><li type="Needs Water"/></sequent>
<sequent><li type="Is Motile"/><entails/><li type="Needs Water"/></sequent>
<sequent>
  <li type="Needs Water"/>
<entails/>
  <li type="Needs Chlorophyll"/>
  <li type="Is Motile"/>
</sequent>
<sequent>
  <li type="Needs Chlorophyll"/>
  <li type="Is Motile"/>
<entails/>
</sequent>
```

New terminology can be introduced with type definitions. Named formal concepts are called *conceptual views*. Conceptual views are type definitions that are restricted to the lattice operations of meet and join. In general, conceptual knowledge is represented by: (1) the three partially ordered sets of objects, attributes and views; (2) the membership or instantiation relation between objects and views whose columns record the *extent* of all the conceptual views; (3) the abstraction relation between views and attributes whose rows record the *intent* of all the conceptual views; and (4) the *incidence* (*classification*) relation between objects and attributes. This aggregate of conceptual knowledge is represented by the notion of a *concept space*.

### THE PROCESS OF CONCEPTUAL SCALING

The application of facets in the theory of library classification was first tested and developed by Ranganathan in his Colon classification system. Recognizing that a classification had to be able to grow organically, in order to keep up with an always-expanding growth in knowledge, the

Colon classification system breaks knowledge down into broad classes, which are further subdivided into single dimensions of information called "facets." Faceted analysis and classification provides a flexible means to classify complex, multi-concept subjects [9]. Complex subjects are divided into their component, single-concept subjects. Single-concept subjects are called *isolates*. Faceted analysis examines the literature of an area of knowledge and identifies its isolates. A *facet* is the sum total of isolates formed by the division of a subject by one characteristic of division. Some examples of facets in musical literature are: composer, instrument, form, etc. Isolates within facets are known as foci.

Comparing these ideas to ontologically structured metadata, facets are identified with conceptual scales or linguistic variables and are often associated with a composite description function, and isolate/foci are identified with scale attributes or linguistic values. A composite description function in the ontology may consist of one function with primitive image values, or a binary relation connecting two types of objects composable with such a function, or something more complex. In conceptual knowledge processing each facet is computed by a conceptual scale. A conceptual scale is an active filter or lens through which information is interpreted. Faceted analysis is conceptual scaling. It involves four steps.

1. Gather ontologically structured metadata.
2. Identify conceptual scales of interest and specify attributes within them (abstract conceptual scale).
3. Specify the structure of conceptual scales (concrete conceptual scale).
4. Apply the conceptual scales to the metadata, producing a composable vector of facets (realized conceptual scales) that constitutes the conceptual space.

The operation of apposition [11], as illustrated by the Intel Press Release conceptual space in Figure 4, combines concept lattices or formal contexts, which share a common collection of objects. This is a pasting or gluing operation with respect to formal contexts: just line up their matrices left to right using a disjoint union of the attribute sets. Apposition is used in conceptual scaling in order to combine dimensions of data. This is the basis for the faceted structure of conceptual space.

**Figure 3: Conceptual Scaling**

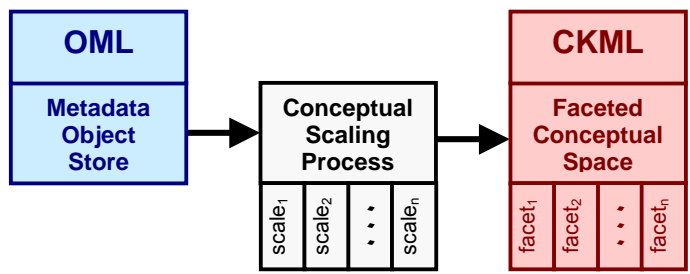

Conceptual knowledge processing is a principled approach to the organization of knowledge. Use of previously developed ideas and techniques, such as apposition and conceptual scaling, offer a decided advantage over more ad hoc approaches for conceptualizing information. Conceptual scaling potentially allows for client-side customization of the organization, and apposition enables transparent composition of content and structure. Conceptual scales are structured by type, kind and method. As depicted in Figure 3, the languages OML and its extension CKML are distinguished by and connected through the process of conceptual scaling.

## Types of Conceptual Scales

The three types of conceptual scales are: abstract, concrete and realized. These types are arrayed along an intentional-extensional dimension. Abstract conceptual scales introduce terms (attribute names), and abstractly specify attribute definitions via term-to-term relationships, as structured by a concept lattice. This lattice is defined, specifically by a set of term implications and term disjointnesses, and generically by propositional sequents. Concrete conceptual scales attach meaning and provide definition to the terms in abstract scales by binding a (single-variable) query to each term. These queries are required to respect the abstract term relationships (implications and disjointnesses). Concrete conceptual scales are conceptual filters or lenses. When applied to a collection of objects (ontologically structured metadata), concrete conceptual scales produce facets, which are known in conceptual knowledge processing as realized scales. Each facet is a concept space representing a single dimension of the metadata. A faceted conceptual space is a vector of facets, and conversely, facets are the components that underlie a faceted conceptual space.

**Table 1: Representations for Types of Conceptual Scales**

| CKP Scale Type | CKML Element |
|----------------|--------------|
| abstract Scale | theory |
| concrete Scale | interpretation |
| realized Scale | local logic |

## Kinds of Conceptual Scales

The process of conceptual scaling is of several *kinds* [3,6,8]: nominal, ordinal, hierarchical, etc. The most common kinds are "nominal" and "ordinal". Nominal scaling refers to names that are members of an unordered set, whereas ordinal scaling refers to subranges of a totally ordered set. Conceptual scaling kind is concentrated in the arguments of the various relations that are associated with the scale. Table 1 describes the various kinds of conceptual scales according to mathematical structure and purpose or use [3]. Mathematical types of scales represent intuitive ideas of design.

**Table 2: Kinds of Conceptual Scales**

| Kind | Mathematical Structure | Purpose/Use |
|------|------------------------|-------------|
| nominal | set | partition/separateness |
| ordinal | (often total) order | ranking |
| interordinal | partial order of intervals | betweenness |
| hierarchical | tree structure | nesting |
| metrical | generalized metric space | similarity |

## Methods of Conceptual Scaling

From the practice of conceptual scaling come several methods of conceptual scaling. These pragmatic methods are closely, but not completely, related to the mathematical structure. They are based primarily upon pragmatic considerations in the practice of conceptual scaling.

- **Direct Scaling:** The scales involved are always realized conceptual scales (facets), and usually are part of the specification of a specific ontology, either a theory [1] in that ontology or a collection (of attributes) controlled by that ontology. They are manifested as binary ontological relation instances, whose first argument is a particular instance (object) of a category (object type) in the generic ontology being specified, and whose second argument is a value in a controlled vocabulary. Examples include: the `genre` relation for movies, the `keyword` relation for press releases, etc. Elements of Dublin Core fit here.

- **Simple Scaling:** This is the classic method of conceptual scaling. Here two ingredients combine to form the state descriptions (rows) of the classification tables: a previously specified domain conceptual scale, either for a datatype or a controlled vocabulary; and a description function from an ontological category to this domain scale. The description function corresponds to a column in a database table. This can be the participant part of the bottom-up relational scaling discussed below. Examples include: time period scaled ordinally or interordinally, person's age scale, etc. Equivalence scaling is a special case.

- **Relational Scaling:** Ontological relations and finite information channels are intimately involved with each other. On the one hand, underlying each finite information channel

$$\mathsf{C} = \{\, f_i : A_i \to C \,\}_{i \in I}$$

where $I = \{1,2,...,n\}$, is an ontological relation, whose relational type is the type set of the channel core, whose relational instances are the channel connections (tokens in the channel core), whose participants (argument domains) are the channel components, and whose argument functions are the token maps underlying the channel infomorphisms. On the other hand, any ontological relation

$$R(A_1, A_2, \ldots, A_n)$$

can be conceptually scaled by viewing it as an information channel.

## Example: Intel Press Releases

Here an introduction to the pragmatics of CKML is made with a real world example - the online press releases of Intel Corporation. This example illustrates the process for conceptualizing information. Here we have gathered and extracted metadata for a series of around 500 press releases ranging over several years. This metadata has been conceptually scaled along four dimensions of information: object type hierarchy, release date order, reference and keyword. These four dimensions, called facets, are the building blocks for the press release conceptual space.

In order to promote good customer relations, many companies enhance their on-line presence with the addition of user interest profiles in either a push or pull modality. Intel has a facility in the pull mode called Custom News, which constructs user interest profiles. On request via a user's bookmark, by applying the user interest profile to Intel site information, a response is built in the form of a "personalized Intel news page." So far, a large part of the Custom News document space consists of Intel press releases.

Unfortunately, much as the Intel site in particular and most corporate sites in general, the structure of the user interest profile is a rather ad hoc affair. As a result, it will scale very poorly when a true Intel intranet is being constructed, and will scale even worse when extranets are being set up between the Intel intranet and other collaborating company intranets. The solution to this problem is to use a good conceptualization for both the Intel sites itself and the profile structure. The Intel conceptualization will be specified by an Intel ontology. When suitably conceptually scaled and view structured, the Intel ontology will become a faceted Intel press release conceptual space, and a user interest profile will be represented by a conceptual view definition.

The Intel press release concept space has four core facets: a object type facet as prescribed by the Intel pressroom archive, a release date facet related to release date and release age, a reference facet based upon the ontological reference relation which a robot produces, and a keyword facet also the result of the robot. Although the Intel ISA hierarchy provides an Intel-wide facet, we will not use this facet here (see the note below which discusses an explicit use of this facet), but instead will implicitly use the ISA hierarchy in the definition of standing queries with quantifiers. The full press release concept space will be composed as the apposition of these four individual facets. It has Intel press releases as its objects and the disjoint union (genus subtypes + release date types + Intel entities + keywords) as its attributes.

### Figure 4: Press Release Faceted Conceptual Space

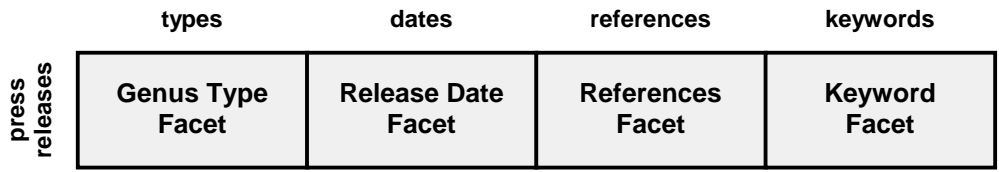

## INSTANCE COLLECTIONS

The Press Release object collection contains metadata for the nearly 500 online Intel press releases. The following markup displays one example of a Press Release object in detail - a press release about the "Infinite CD." In general, the genus category must be situated between the lowest categories within which the object declares membership and the top "Intel Entity" category.

```
<Collection.Object id = "intel-press-releases" genus="Press-Release"
  ontology = "http://www.intel.com/ontology/">
   • • •
  <Press-Release id = "cn022497"
     text = "Intel, Marimba And Macromedia To Create
             Infinite CD For Public Broadcasting Service"
     about = "http://www.intel.com/pressroom/archive/releases/cn022497.htm"
     image = "infiniteCD.gif">
     <comment>
       An Intel Press Release: Technology will enable PBS ONLINE to deliver
       Customized media-enhanced programming and products.
     </comment>
     <classification type = "Corporate News"/>
     <city  target.Instance = "San Francisco"/>
     <state target.Instance = "California"/>
     <date  target.Instance = "1997/02/24"/>
     <language  target.Instance = "English"/>
     <reference target.Instance = "Company#Intel"/>
     <reference target.Instance = "Company#Marimba"/>
     <reference target.Instance = "Company#Macromedia"/>
     <reference target.Instance = "Company#PBS"/>
     <reference target.Instance = "Person#'John Hollar'"/>
     <reference target.Instance = "Person#'Kim Polese'"/>
     <reference target.Instance = "Executive#'Claude Leglise'"/>
     <reference target.Instance = "'Web Site'#'Connected PC'"/>
     <reference target.Instance = "Product#Pentium"/>
     <reference target.Instance = "Product#MMX"/>
     <reference target.Instance = "Product#Intercast"/>
     <keyword target.Instance = "Keyword#multimedia"/>
     <keyword target.Instance = "Keyword#push"/>
     <keyword target.Instance = "Keyword#cd-rom"/>
  </Press-Release>
   • • •
</Collection.Object>
```

## *THEORIES*

The following CKML theories anchored at the Press Release type use only the most basic definition for conceptual scales. Object types defined by complex queries are not used here, although they could have been. The release date theory (conceptual scale) is a complete ordinal scale consisting of half-closed intervals. No closed intervals representing between-ness are used. As a result, the release date facet is a linear (total) order. The reference conceptual scale is partitioned into three subscales based upon subtypes of the root Intel Entity type: Executive, Product and Company. Each subscale is a complete nominal conceptual scale. As with all other conceptual spaces, the genus type facet is defined by the implicit hierarchical conceptual scale consisting of all object types at or below the Press Release type, the genus of the Press Release theory.

```
<Theory name = "Release Date" genus = "Press Release">
  <Interpretation name = "date ordinal">
    <Foreach var = "date" type = "Date">
      <Where>
        <subrange var = "date" begin = "1995/02/01" end = "1997/08/01"/>
      </Where>
      <Object var = "pr" type = "Press Release">
        <comment>
          This press release was released on or after the specific date.
        </comment>
        <date order = "geq" source.Instance = "pr" target.Instance = "date"/>
      </Object>
    </Foreach>
  </Interpretation>
</Theory>

<Theory name = "Reference" genus = "Press Release">
  <Interpretation name = "reference nominal">
    <Foreach var = "entity" type = "Executive">
      <Object var = "pr" type = "Press Release">
        <comment>
          This press release references the specific executive.
        </comment>
        <reference source.Instance = "pr" target.Instance = "entity"/>
      </Object>
    </Foreach>
    <Foreach var = "entity" type = "Product">
      <Object var = "pr" type = "Press Release">
        <comment>
          This press release references the specific executive.
        </comment>
        <reference source.Instance = "pr" target.Instance = "entity"/>
      </Object>
    </Foreach>
    <Foreach var = "entity" type = "Company">
      <Object var = "pr" type = "Press Release">
        <comment>
          This press release references the specific executive.
        </comment>
        <reference source.Instance = "pr" target.Instance = "entity"/>
      </Object>
    </Foreach>
  </Interpretation>
</Theory>

<Theory name = "Keyword" genus = "Press Release">
  <Interpretation name = "reference nominal">
    <Foreach var = "keyword" type = "Keyword">
      <Object var = "pr" type = "Press Release">
        <comment>
          This press release uses the specific keyword.
        </comment>
        <reference source.Instance = "pr" target.Instance = "keyword"/>
```

```
        </Object>
      </Foreach>
</Theory>
```

## Interoperability with CKML

Interoperability is very important for a language whose goal is to represent distributed information in a conceptual framework. The discussion in this section demonstrates how CKML is interoperable with many important alternative approaches touted as solutions to all or part of this goal: Resource Description Framework with Schemas (RDF/S), Ontolingua, conceptual graphs as codified in the Conceptual Graph Interchange Form (CGIF), and relational and object-oriented databases. Each of these is discussed in the following subsections.

### RESOURCE DESCRIPTION FRAMEWORK WITH SCHEMAS (RDF/S)

RDF/Schemas has the structure of a semantic network. It corresponds to simple conceptual graphs [10], which are conceptual graphs without negations, universal quantifiers and nested conceptual contexts. To support metadata interoperability, the part of OML that corresponds to RDF with Schemas (expressions with no negations, implications, conceptual contexts, or universal quantifiers) will be demarcated as Simple OML.

At the core of Simple OML, OML and CKML is a classification-projection semantics. This classification-projection semantics, which is visualized in Diagram 3, asserts the axiom that "*for any binary relational instance r and binary relational type t, if t is a binary relational type of r then the source (target) type of t is an entity type of the source (target) entity of r.*" Symbolically

$$r \vDash t \text{ implies both } \partial_0(\,r\,) \vDash \partial_0(\,t\,) \text{ and } \partial_1(\,r\,) \vDash \partial_1(\,t\,).$$

All type names in Diagram 3 denote higher order types (meta-types). The Entity type partitions

**Entity** = **Object + Data**

into the Object type and the Data type. Here the plus denotes disjoint union or type sum. So data values are on a par with object instances, although of course less complex. The Data type includes all primitive types and enumerated types.

**Diagram 3: Classification Projection**

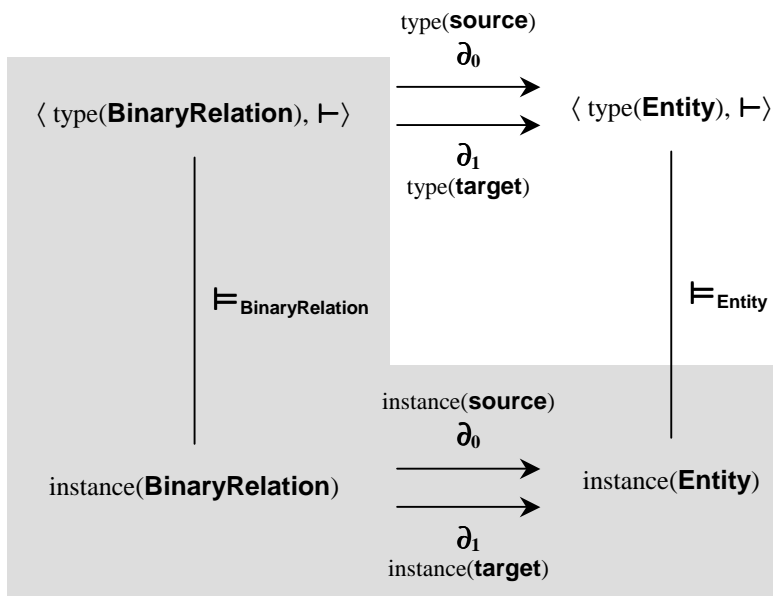

The top subdiagram of Diagram 3 owes much to category theory and type theory. A category is defined to be a collection of objects and a collection of morphisms (arrows), which are connected by two functions called source (domain) and target (codomain). To complete the picture the composition and identity operators need to be added, along with suitable axioms. The partial orders on objects and arrows represent the type order on entities and binary relations. The bottom subdiagram extension gives a pointed version of category theory, a subject closely related to elementary topos theory. The classification relation connects the bottom subdiagram (instances) to the top subdiagram (types), and represents the classification relation of Barwise's Information Flow [1]. The shaded part of Diagram 3 corresponds to RDF without classes. Elements of this correspondence are listed in Table 3. The question mark in Table 3 reflects the current undeveloped state of RDF/S data types. These are being developed by the XML Schema working group of the W3C, and will be incorporated into CKML when finalized.

**Table 3: RDF/S and Simple OML Correspondences**

| RDF/S notion | Simple OML notion | formalism |
|---|---|---|
| class | object type | type(**Object**) |
| ???? | data type | type(**Data**) |
| property | binary relation type | type(**BinaryRelation**) |
| subClassOf | subtype on objects | $\vdash$**Entity** |
| subPropertyOf | subtype on binary relations | $\vdash$**BinaryRelation** |
| domain | type source | type(**source**) $= \partial_0$ |
| range | type target | type(**target**) $= \partial_1$ |
| resource | object instance | instance(**Object**) |
| literal | data type value | instance(**Data**) |
| statement | binary relation instance | instance(**BinaryRelation**) |
| subject | instance source | instance(**source**) $= \partial_0$ |
| object | instance target | instance(**target**) $= \partial_1$ |
| predicate, type | classification | $\vdash$**BinaryRelation**, $\vdash$**Entity** |

The fact that this simple structure for OML corresponds closely to the core structure of RDF/S, illustrates why the core part of the RDF/S syntax is embeddable into the Simple OML syntax. The Simple OML serialization syntax is the closest approach to the RDF/S serialization syntax. The most obvious difference is the lack of types in basic RDF - these are to be modeled with schemas. Types are not considered as essential in RDF as they are in OML/CKML, since schema classes are just special kinds of RDF resources. This is reasonable and is close to the frame system approach, but it is different from the conceptual framework of OML/CKML, which is based more on Sowa's conceptual graphs [10] (see Table 5) and particularly Barwise's theory of information flow [1]. Although RDF Schema classes are normally modeled as types, in order to model the RDF semantics that "properties are resources," they could be modeled in OML/CKML as special objects, with explicit models for the subclass partial order relation between classes, the classification relation between resources and classes, the domain and range functions, etc.

OML allows both free and embedded binary relations, whereas RDF properties are only embedded. The OML namespace mechanism is a bit different from the RDF namespace mechanism: any real-world object is represented by an OML object (surrogate) with a link to the real-world object and OML references to the real-world object are made through this surrogate, whereas web resources may be referenced in RDF without being described (represented); and the complete references for an OML object (instance) has the 3-fold syntax *ontology* : *type* # *identifier*. This is an extension of the [XML namespace mechanism](). Finally, OML collections are typed, in contrast with RDF bags, which are not.

## Quotations: Statements about Statements

Quotations, which are statements about statements, are a basic component of Tim Berners-Lee's notion of a [semantic web](). The RDF model for quotations needs to use reification. In contrast, OML/CKML has two models for quotations: one using reification like RDF, and the other using *conceptual contexts*, an idea from conceptual graphs. The conceptual context model for quotations needs to use assertions, and the ability to nest assertions within objects using the description attribute. Being a complete semantic model for conceptual graphs, OML/CKML has conceptual contexts. RDF should have such a capability, but doesn't, and thereby is impoverished.

### Example: WWW Quotations

Consider the following example from the RDF specification document of a "statement about a statement."

> **"Ralph Swick says that Ora Lassila is the creator of the resource http://www.w3.org/Home/Lassila."**

This fact is expressed as follows in RDF, conceptual graphs and OML.

RDF (Resource Description Framework)

```
<rdf:RDF
  xmlns:rdf= "http://www.w3.org/1999/02/22-rdf-syntax-ns#"
  xmlns:a  = "http://description.org/schema/">

  <rdf:Description>
    <rdf:subject resource = "http://www.w3.org/Home/Lassila"/>
    <rdf:predicate resource = "http://description.org/schema/Creator" />
    <rdf:object>Ora Lassila</rdf:object>
    <rdf:type resource = "http://www.w3.org/1999/02/22-rdf-syntax-ns#Statement" />
    <a:attributedTo>Ralph Swick</a:attributedTo>
  </rdf:Description>

</rdf:RDF>
```

Conceptual Graph Display Form

The proposition concept contains a conceptual graph in its referent field. This is an example of a *context* in conceptual graphs.

**Diagram 4: Conceptual Context**

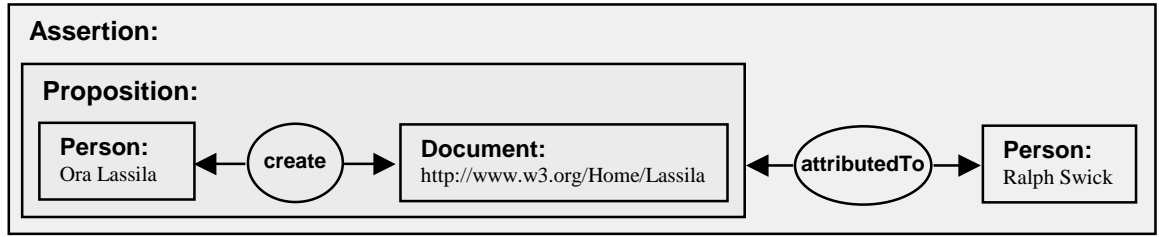

OML (Ontology Markup Language)

There are two ways of doing this: using reification, or using conceptual contexts.

*Reification*

```
/* types in an ontology */
<Type.Object name = "Document"/>
<Type.Object name = "Person"/>
<Type.BinaryRelation name = "create"
  source.Type = "Person" target.Type = "Document"/>
<Type.BinaryRelation name = "attributedTo"
  source.Type = "Proposition" target.Type = "Person"/>
<Type.Relation name = "Create" binrel = "create">        /* reified relation */
  <Type.Function name = "agent" target.Type = "Person"/>
  <Type.Function name = "theme" target.Type = "Document"/>
</Type.Relation>

/* instances in a collection */
<Document id = "ora-homepage" about = "http://www.w3.org/Home/Lassila"/>
<Person id = "ora" text = "Ora Lassila"/>
<Person id = "ralph" text = "Ralph Swick"/>
<Instance.ReifiedBinaryRelation id = "ora-create-homepage"
  source.Instance = "ora-home-page"                      /* subject   */
  target.Instance = "ora"/>                               /* object    */
  <classification type = "creator"/>                      /* predicate */
  <attributedTo target.Instance = "ralph"/>
</Instance.ReifiedBinaryRelation >
```

*Conceptual Context*

The part that is in OML, but not in Simple OML, is emphasized. Note how the description attribute for the Proposition instance defines a propositional *context*, by pointing to the contents of the context – the assertion.

```
/* types in an ontology */
<Type.Object name = "Document"/>
<Type.Object name = "Person"/>
<Type.Object name = "Proposition"/>
<Type.BinaryRelation name = "create"
  source.Type = "Person" target.Type = "Document"/>
<Type.BinaryRelation name = "attributedTo"
  source.Type = "Proposition" target.Type = "Person"/>
<Type.Relation name = "Create" binrel = "create">        /* reified relation */
  <Type.Function name = "agent" target.Type = "Person"/>
  <Type.Function name = "theme" target.Type = "Document"/>
</Type.Relation>
```

```
/* instances in a collection */
<Document id = "ora-homepage" about = "http://www.w3.org/Home/Lassila"/>
<Person id = "ora" text = "Ora Lassila"/>
<Person id = "ralph" text = "Ralph Swick"/>
<Proposition id = "ora-create-homepage" description = "assertion-create"/>
<Assertion id = "assertion-create">
  <create source.Instance = "ora" target.Instance = "ora-homepage">
</Assertion>
<attributedTo source.Instance = "ora-create-homepage" target.Instance = "ralph"/>
```

### Reification

According to Webster, to *reify* means "to regard (something abstract) as a material or concrete thing." According to Winston [12], reification means "treating something abstract and difficult to talk about as though it were concrete and easy to talk about." Winston's technical definition of reification is the elevation of a link in a semantic network to the status of a describable node.

When reification of binary relations is done correctly, functions are needed. Reification introduces function types called *participants* or *thematic roles* of the original relation. As the example above illustrates, although the original relation might be binary, the reified relation may have additional participants defined. Some canonical participants are: the agent, a coagent, the beneficiary, the thematic object, the instrument, the time of the event, the location of the event, etc. The agent is the entity that causes the action. A coagent serves as partner to the principal agent. The beneficiary is the entity for whom the action is performed. The thematic object is the entity with which the relationship is concerned. The instrument is the tool used by the agent to realize the relationship.

CKML has two forms of reification: an implicit reification represented by the relation-object type ordering, and an explicit reification for binary relations. In describing the OML/CKML version of reification, all reified binary relations are instances of the `CKML:Relation` type, a subtype of the `CKML:Object` type. The fact that a multivalent relation is a special case of an object in CKML, represented by the fact that `CKML:Relation` type is a subtype of the `CKML:Object` type, is a form of reification that is built into OML/CKML.

Two apparent reasons for reification are: to analyze a statement in order to discover the thematic roles played by various objects (represented by noun phrases), or to make a statement about another statement (use a quotation). As pointed out by the rendering of the above example in OML/CKML, reification is not necessary for this. Instead, it may be better to model quotations using the idea of a conceptual context. The use of reification to model quotations may confuse the expression of the propositional content of one statement with a second statement that references the first statement. An automatic translator might have trouble separating the expressions of the two statements. The expression of the propositional content of a statement can be made in OML /CKML with an assertion (equivalent to a conceptual graph). For conceptual contexts, we need both relations and assertions in order to adequately express "statements about statements". For Simple OML we can ignore relations, and just use the supertype of objects. Apparently however, we cannot ignore assertions. Using contexts to model quotations requires full OML language.

### *ONTOLINGUA*

Ontolingua and OML/CKML were developed independently, and it is remarkable how similar they are. The table below shows the correspondences between Ontolingua and OML/CKML.

**Table 4: Ontolingua-CKML Correspondences**

| Ontolingua | OML/CKML |
|---|---|
| Class | Object type |
| Relation | Relation, BinaryRelation |
| Function | Function |
| Individual | Instance |
| Axiom | Assertion |
| => | <implies> |
| <=> | <equiv> |
| and | <and> |
| Or | <or> |
| Not | <not> |
| Exist | <Exists> |
| | <Forall> |
| ?someVar | var = "someVar" |
| Frame | Object type, Function, Assertion |
| Slot | BinaryRelation, Function |
| " ... " | <comment> |

An example from the glossary of ontology terminology in the Ontolingua tutorial illustrates the use of these correspondences in the axiomatics of Ontolingua and OML/CKML.

Natural Language

*"If any two animals are siblings, then there exists someone who is the mother of both of them."*

Ontolingua Axiom

```
((?sib1 Animal)
 (?sib2 Animal))
(=>
  (sibling ?sib1 ?sib2)
  (exists ((?mom Animal))
    (and
      (has-mother ?sib1 ?mom)
      (has-mother ?sib2 ?mom)
    )
  )
)
```

OML/CKML Assertion

```
<Forall var = "sib1" type = "Animal">
<Forall var = "sib2" type = "Animal">
  <implies>
    <sibling source.Instance = "sib1" target.Instance = "sib2">
    <Exists var = "mom" type = "Animal">
      <and>
        <has-mother source.Instance = "sib1" target.Instance = "mom">
        <has-mother source.Instance = "sib2" target.Instance = "mom">
```

```
        </and>
      </Exists>
    </implies>
  </Forall>
</Forall>
```

There are differences however. In particular, OML/CKML is strictly a two-tiered structure between types and instances (IF tokens), and does not have a correspondent to the Ontolingua instance. And a related idea is that Ontolingua classes-of-classes often corresponds to OML/CKML theories.

**Figure 5: CKML Vehicle Ontology**

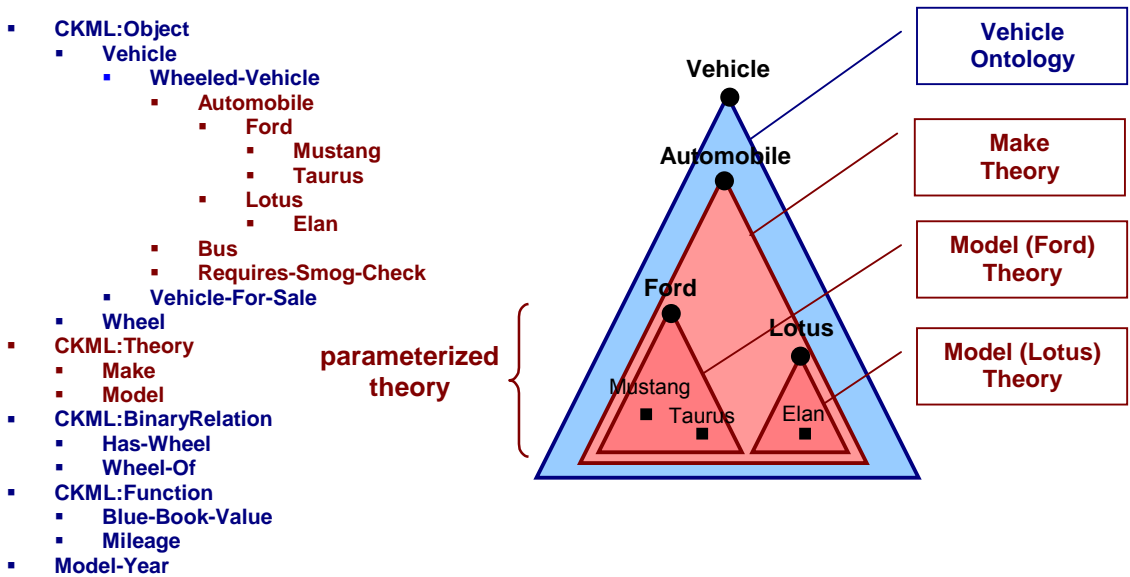

- **CKML:Object**
  - **Vehicle**
    - **Wheeled-Vehicle**
      - **Automobile**
        - **Ford**
          - **Mustang**
          - **Taurus**
        - **Lotus**
          - **Elan**
      - **Bus**
      - **Requires-Smog-Check**
    - **Vehicle-For-Sale**
    - **Wheel**
- **CKML:Theory**
  - **Make**
  - **Model**
- **CKML:BinaryRelation**
  - **Has-Wheel**
  - **Wheel-Of**
- **CKML:Function**
  - **Blue-Book-Value**
  - **Mileage**
- **Model-Year**

The hierarchy in Figure 5 gives the OML/CKML representation of the Vehicle example in the Ontolingua online tutorial.

## CONCEPTUAL GRAPHS (CGIF)

**Table 5: CGIF-CKML Correspondences**

| CGIF | OML/CKML |
|------|----------|
| Conceptual Graph | Assertion element |
| Concept | object instance |
| Conceptual Relation | relation instance |
| Lambda Expression | Lambda element |
| Concept Type | Object element |
| Relation Type | Relation element |
| Referent | ••• |
| Quantifier | Exists, Forall elements |
| Designator | ••• |
| Literal | multimedia attributes |
| Locator | about attribute |

| Conceptual Graph | Expression element |
|---|---|
| Context | Assertion element<br>referenced by<br>description attribute in object instance |
| Coreference Set | ••• |
| defining label | id attribute (namespace) |
| bound labels | obj attributes (namespace) |
| Knowledge Base | OML element (bracket) |
| Type hierarchy | Ontology element |
| Relation hierarchy | Ontology element |
| Catalog of individuals | Collection elements (object kind) |
| Outermost context | Collection elements (relation kind) |

OML has an abbreviated form that, among other things, replaces lambda/quantifier brackets with lambda/quantifier attributes, eliminates function and argument elements, and replaces the function-star expressions in object type definitions by arbitrary expressions in the knowledge base catalogue of individuals. A grammar for Abbreviated OML is now under development. We have recommended to the conceptual graph community that the abbreviated form of OML serve as a base for the development of a Conceptual Graph Markup Language.

### DATABASE REPRESENTATION

**Figure 6: Database Ontology Hierarchy**

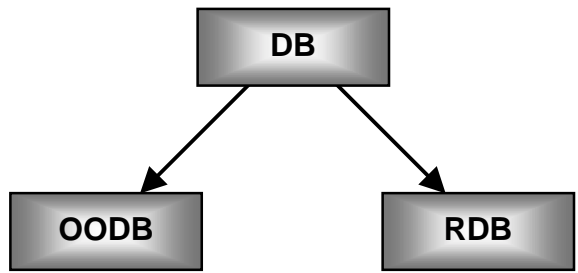

CKML can be used to interrelate different kinds of databases. Relational databases are relation-centric and represent information in tables, whereas object-oriented databases are object-centric. Either kind of database can be translated to the other by suitable definitions in CKML. But such a translation is best defined in terms of a generic database ontology. As shown in the ontology extension hierarchy of Figure 6, in this CKML database representation there are three ontologies representing database schemas: a generic database ontology (DB), a specific object-oriented database ontology (OODB), and a specific relational database ontology (RDB). The arrows in the figure represent both type inclusion and type equivalence (synonymy). All the additional types in the object-oriented ontology and the relational ontology are defined in terms of the basic types in the generic ontology.

**Figure 7: Block Structures**

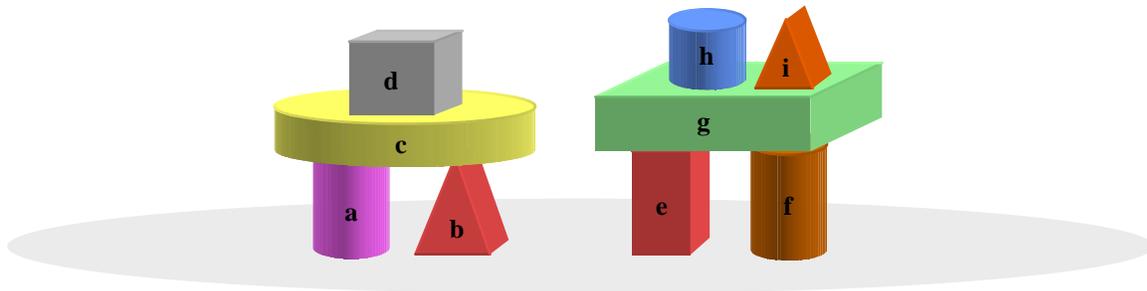

## Example: Database Connectivity with CKML

To illustrate how CKML is related to the knowledge represented in databases, we expand on a conceptual graph example from chapter 7 in the book *Knowledge Representation* by John Sowa [10] and rephrase it in terms of CKML ontologies and instance collections. This example, called *block structures,* consists of various kinds of blocks stacked on top of one another as in Figure 7.

### Generic Database

The basic entity types are: Block, Color, Shape and Support. The basic binary relation and function types are: shape, color, thme and inst. Shape and Color, which represent database domains, are specified in CKML as data type definitions. The Support object type is reified from a support binary relation type. There is a generic classification relation type chrc, called *characteristic*, which connects object instances with higher order object types. The block notion is described by the two basic function types, shape and color. These are restrictions of chrc to the entity subtypes Shape and Color. The support notion is described by the two basic function types, inst and thme. These are restrictions of two generic participation relation types, inst and thme, called *instrument* and *theme* (or *thematic object*), which are used in the representation of the semantics of natural language.

The basic types are specified in the following generic ontology.

(located at http://www.database.org/ontology/db/)

```
<Ontology name="Generic Database">
  <Type.Object name= "Block">
    <Type.Function name="shape" target.Type="Shape"/>
    <Type.Function name="color" target.Type="Color"/>
  </Type.Object>
  <Type.Data name="Shape" ordered="no">
    <value name="cubical"/>
    <value name="prismatic"/>
    <value name="pyramidical"/>
    <value name="cylindrical"/>
    <value name="conical"/>
    <value name="spherical"/>
  </Type.Data>
  <Type.Data name="Color" ordered="no">
    <value name="red"/>
    <value name="orange"/>
    <value name="yellow"/>
    <value name="blue"/>
    <value name="indigo"/>
    <value name="violet"/>
    <value name="brown"/>
    <value name="gray"/>
    <value name="white"/>
    <value name="black"/>
  </Type.Data>
  <Type.BinaryRelation name="support"
    source.Type="Block" target.Type="Block"/>
  <Type.Object name="Support">  /* reified relation */
    <Type.Function name="inst" target.Type="Block"/>
    <Type.Function name="thme" target.Type="Block"/>
  </Type.Object>
</Ontology>
```

*Object-Oriented Database*

The additional object-oriented database types define new terminology for blocks. They are specified in terms of the basic types in the generic database ontology. These additional types, which are subtypes of the Block type, are defined in a theory interpretation via role restriction. This theory interpretation is an example of a concrete conceptual scale defined by the simple nominal conceptual scaling of the basic shape function.

(located at http://www.database.org/ontology/oodb/)

```
<Ontology name="Object-Oriented Database">
  <extends ontology="http://www.database.org/ontology/db/" prefix="DB"/>
  <Interpretation function.Type="DB:shape">
    <Type.Object name="Pyramid" var= "x" type="DB:Block">
      <DB:Block id="x" shape="DB:Shape#pyramidical"/>
    </Type.Object>
    <Type.Object name="Cylinder" var= "x" type="DB:Block">
      <DB:Block id="x" shape="DB:Shape#cylindrical"/>
    </Type.Object>
    ...
  </Interpretation>
  <Type.Set name="Set.Block" genus="DB:Block"/>
  <Type.Function name="support"
    source.Type="DB:Block" target.Type="Set.Block"/>
  <Type.Collection name="Collection.OODB" genus="CKML:Object"/>
</Ontology>
```

There is an implicit theory behind the theory interpretation in the object-oriented database ontology, which puts no constraints on the subtypes. Constraints, such as partition of the genus type Block, could be specified a priori, in a more generic ontology, as the following explicit theory. This theory represents an abstract conceptual scale specifying a set of constraints that the concrete conceptual scale, represented by the theory interpretation, must satisfy.

```
  <Theory name="Block" genus="DB:Block">
    <Type.Object name="Pyramid"/>
    <Type.Object name="Cylinder"/>
    ...
    <partition>
      <li type="Pyramid"/>
      <li type="Cylinder"/>
    ...
    </partition>
  </Theory>
```

Diagram 4 is the representation of (part of ) the block structures of Figure 1 as an object-oriented database. Objects correspond to blocks. The set-valued support function is used.

The following table lists the CKML representation of the block structures as an object-oriented collection in specific style with function abbreviation.

```
<Collection.OODB ontology= "http://www.blockstructures.org/ontology/oodb/">
  <Cylinder id="a" color="Color#violet">
    <support><Set.Block><li instance="c"/></Set.Block></support>
  </Cylinder>
  <Pyramid id="b" color="Color#red">
    <support><Set.Block><li instance="c"/></Set.Block></support>
  </Pyramid>
  ...
  <Cylinder id="f" color="Color#brown">
    <support><Set.Block><li instance= "g"/></Set.Block></support>
  </Cylinder>
```

```
    <Prism id="g" color="Color#green">
      <support><Set.Block>
        <li instance="h"/>
        <li instance="h"/>
      </Set.Block></support>
    </Prism>
    ...
</Collection.OODB>
```

**Diagram 4: Object-oriented Database (part)**

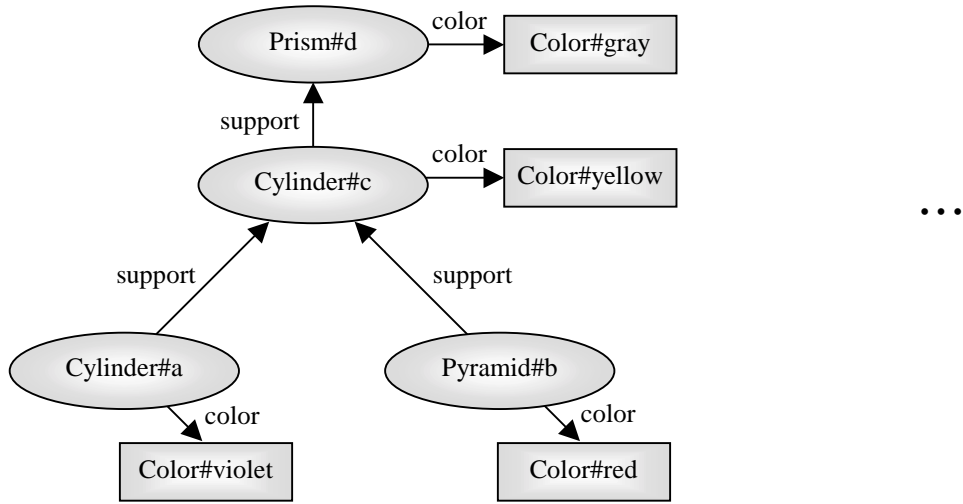

*Relational Database*

The relational database types (for domains and tables) are specified in terms of the basic types in a specific relational database ontology that extends the generic database ontology. The two basic types in the CKML relational database representation are Object and BinaryRelation. A multivalent relation is represented by the Relation type, a subtype of the Object type. The multivalent relations of conceptual graphs fit the relational database model precisely, since they do not name participants (participating arguments).

**Table 6: Correspondences between CG and RDB**

| CG | RDB |
|---|---|
| relation | relation (table) |
| concept | domain (table column) |
| individual marker or literal | value (table row or entry) |

Except for the binary relation, CKML does not fit the relational database model as precisely, since, except for binary relations, participants are named (see query processing below) in CKML. The deficit lies in the relational database model, which represents participation only indirectly through use of domain names. On the other hand, CKML fits the object model better than conceptual graphs, for the same reason.

(located at http://www.database.org/ontology/rdb/)

```
<Ontology name="Relational Database">
  <extends ontology="http://www.database.org/ontology/generic/" prefix="DB"/>
  <Type.Object name="Supporter" var="x" type="OO:Block">
    <OO:Support inst="x"/>
  </Type.Object>
  <Type.Object name="Supportee" var="x" type="OO:Block">
    <OO:Support thme="x"/>
  </Type.Object>
  <Type.BinaryRelation name="support"        /* reverse reification */
    source="y" source.Type="Supporter"
    target="z" target.Type="Supportee">
      <DB:Support inst="Block#y" thme="Block#z"/>
  </Type.BinaryRelation>
  <Type.Collection name="Collection.Block" genus="DB:Block"/>
  <Type.Collection name="Collection.Support" genus="DB:Support"/>
</Ontology>
```

Table 7 is the representation of the block structures of Figure 7 as a relational database. There are two database relations: the Block relation lists each block's ID (identifier), plus its shape and color (dimensional data is ignored); the Support relation lists the immediate block supporting relationships.

**Table 7: Block Structures Relational Database**

*Block*

| ID | Shape | Color |
|----|-------|-------|
| a | cylindrical | violet |
| b | pyramidical | red |
| c | cylindrical | yellow |
| d | prismatic | gray |
| e | prismatic | red |
| f | cylindrical | brown |
| g | prismatic | green |
| h | cylindrical | blue |
| i | pyramidical | orange |

*Support*

| Supporter | Supportee |
|-----------|-----------|
| a | c |
| b | c |
| c | d |
| e | g |
| f | g |
| g | h |
| g | i |

The next two tables list the CKML representation of the block structures as a relation-oriented collection. The first collection corresponds to the Block database entity type, and the second collection corresponds to the Support database relation type.

```
<Collection.Block ontology="http://www.blockstructures.org/ontology/rdb/">
  <Block id="a" shape="Shape#cylindrical" color="Color#violet">
  <Block id="b" shape="Shape#pyramidical" color="Color#red">
  ...
</Collection.Block>
```

```
<Collection.Support ontology="http://www.blockstructures.org/ontology/rdb/">
  <support source.Instance="a" target.Instance="c"/>
  <support source.Instance="b" target.Instance="c"/>
  <support source.Instance="c" target.Instance="d/>
  ...
</Collection.Support>
```

## Query Processing

Database queries can be represented as CKML expressions with question marks in place of variables. To access a database, either object-oriented or relational, the type definitions can be used to translate the types in the query expression to matching types in the database. In this case note that the following types are inter-translatable.

**Table 8: Relation Type Equivalences**

|  | Type | Source (domain) | Target (range) |
|---|---|---|---|
| **Object** | DB:Support | inst → DB:Block | thme → DB:Block |
| **Binary Relation** | DB:support | source.Instance → DB:Block | target.Instance → DB:Block |
| **Set-valued Function** | OODB:support | source.Instance → DB:Block | target.Instance → OODB:Set.Block |
| **Binary Relation** | RDB:support | source.Instance → DB:Block | target.Instance → DB:Block |

As a query-processing example, consider the natural language query "Which prism is supported by a cylinder?". This can be expressed as the following CKML query expression.

```
<DB:Support inst="Cylinder" thme="Pyramid#?/>
```

which is an abbreviation for the generic database expression

```
<Exists var="x" type="OODB:Cylinder">
   <DB:Support inst="x" thme="OODB:Pyramid#?/>
</Exists>
```

Translating (reverse reification) from the Support object type to the support relation type results in the expression

```
<Exists var="x" type="OODB:Cylinder">
   <OODB:Cylinder id="x">
      <DB:support target.Instance="OODB:Pyramid#?"/>
   </OODB:Cylinder>
</Exists>
```

The answer to the query, that "prism g is (directly) supported by cylinder f," can be derived by matching this (ignoring the color attribute) with the expression

```
<OODB:Cylinder id="f" color="Color#brown">
   <OODB:support>
      <OODB:Set.Block><li instance= "g"/></OODB:Set.Block>
   </OODB:support>
</OODB:Cylinder>
```

in the object-oriented database that is expressed with the support set-valued function.

To access a relational database such as Table 7, the definitions can be used to translate the types in the CKML query expression. This results in the relational database expression

```
<RDB:support source.Instance="Supporter#x" target.Instance="Supportee#?">
<DB:Block id="x" shape="Shape#cylindrical"/>
<DB:Block id="?" shape="Shape#prismatic"/>
```

which can be translated to the following SQL expression

```
SELECT Supportee
   FROM support, Block x, Block y
      WHERE
            Supporter = x.ID AND Supportee = y.ID
      AND   x.Shape = 'cylindrical'
      AND   y.Shape = 'prismatic'
```

The query-to-SQL translation rules are as follows.

1. Write binary relations in non-embedded form.
2. Use type qualifiers in instance namespace names.
3. Map the question mark to the SELECT part of the SQL query.
4. List the binary relation and the two objects (with identifiers) in the FROM part of the SQL query.
5. List the references between participants as constraints in the WHERE part of the SQL query.

## Evaluation of OML/CKML

The development, representation and sharing of ontological information is very important in bioinformatics and the molecular biology community. Indeed, every biological database employs an ontology, either implicitly or explicitly, to model its data. Two recent workshops ISMB98 - Ontologies for Molecular Biology and ISMB99 - Bio-Ontologies '99 and an associated working group The Bio-Ontology Consortium have been involved in an initiative to evaluate a number of alternative ontology-exchange languages, and to recommend one or more languages for use within the larger bioinformatics community. The evaluation effort involved three separate meetings in 1998 and 1999 by the authors, as well as experiments with the proposed ontology languages.

In phase I of the evaluation, the authors selected a set of candidate languages, and a set of capabilities that the ideal ontology-exchange language should satisfy. Nine candidate languages were evaluated in this phase.

1. **Ontolingua:** developed at Stanford University for the exchange of ontologies; based upon Knowledge Interchange Format (KIF) and the frame knowledge representation systems developed by knowledge-representation researchers.
2. **CycL:** the underlying representation language for the Cyc knowledge representation system; developed at Microelectronics & Computer Technology Corporation (MCC) and Cycorp.
3. **OML/CKML:** a new language pair for representing ontologies, conceptual knowledge and distributed logical information; based upon conceptual graphs, formal concept analysis and information flow [1].
4. **OPM (Object-Protocol Model):** a product from GeneLogic; used in a number of Pharmaceutical/Biotech organizations; uses an underlying object-oriented federated schema for the integration of multiple information sources.
5. **XML/RDF:** being developed by the W3C (World Wide Web Consortium); intended to encode metadata concerning web documents; XML is currently the leading candidate for a generic language for the exchange of semi-structured objects.
6. **UML (Unified Modeling Language):** a set of notational conventions used by software application designers/developers to model their software system; developed by Rational Software and currently backed by Rational, Microsoft and the OMG (Object Management Group).
7. **OKBC (Open Knowledge Base Connectivity):** an application programmers interface (API) for accessing and modifying multiple, heterogeneous knowledge bases; not actually an ontology exchange language; its knowledge model is designed to capture ontologies; successor of Generic Frame Protocol (GFP), a frame representation system developed at the Artificial Intelligence Center at SRI
8. **ASN.1 (Abstract Syntax Notation):** has historical significance as an early language for the exchange of datatypes and simple objects; used in a number of bioinformatics applications from the NCBI (National Center for Biotechnology Information); used in conjunction with the Unified Medical Language System (UMLS) project at the National Library of Medicine (NLM).
9. **ODL (Object Definition Language)**: a relatively new standard coming out of the Object Database Management Group (ODMG); currently a de facto standard for a common representation of objects for object-oriented databases and programming languages.

As a result of the first phase evaluation two languages, Ontolingua and OML/CKML, were selected for further in-depth evaluation in the second phase.

The second phase of the evaluation process focused on the two candidate languages that were deemed most interesting from the initial evaluation: Ontolingua and OML/CKML. The consortium decided that it would be useful to create a small model in each language in order to judge the utility and the representational richness of each language. Sets of experiments were developed to perform this detailed evaluation involving several bio-ontologies: the Riley ontology, the GeneClinics ontology, the Gene ontology, etc. The results of this evaluation suggest two directions for future work: development of an XML expression for the Ontolingua model, or development of a frame-based version of OML/CKML called Simple OML. The exchange language evaluation process by the bio-ontology consortium is still in progress.

## Summary

This paper has discussed the semantics, syntax and pragmatics of Conceptual Knowledge Markup Language. Future work includes implementation of conceptual scaling with federations of brokering agents. In addition, natural language processing techniques will be utilized in client-side specification of conceptual scales. The latter rests largely upon the remarkable similarity between the notion of named formal concept (conceptual view) and the notion of composite conceptual description [13], a similarity that ties CKML closely to description logics.